# ON THE EFFECTS OF IMAGE QUALITY DEGRADATION ON MINUTIAE- AND RIDGE-BASED AUTOMATIC FINGERPRINT RECOGNITION

*J. Fierrez-Aguilar, L.-M. Muñoz-Serrano, F. Alonso-Fernandez and J. Ortega-Garcia*

ATVS/Biometrics Research Lab., Escuela Politecnica Superior, Avda. Francisco Tomas y Valiente, 11
Universidad Autonoma de Madrid, Campus de Cantoblanco, 28049 Madrid, Spain
email: {julian.fierrez, fernando.alonso, javier.ortega}@uam.es

## ABSTRACT

The effect of image quality degradation on the verification performance of automatic fingerprint recognition is investigated. We study the performance of two fingerprint matchers based on minutiae and ridge information under varying fingerprint image quality. The ridge-based system is found to be more robust to image quality degradation than the minutiae-based system for a number of different image quality criteria.

## 1. INTRODUCTION

The increasing need for reliable automated personal identification in the current networked society has resulted in the popularity of *biometrics* [1]. The challenge of biometrics is to reliably differentiate between different subjects according to some target application based on one or various signals measured from physical and/or behavioral traits such as fingerprint, face, iris, voice, hand, written signature, etc.

Within biometrics, automatic fingerprint recognition [2] motivates great interest mainly because of the widespread deployment of electronic acquisition devices, and the number of practical applications ranging from access control to forensic identification. Although commercial applications exist, and contrary to the common belief, automatic fingerprint recognition is still an open issue [2].

One of the open issues in fingerprint verification is the lack of robustness against image quality degradation [3]. This fact is partially corroborated by the results of the last International Fingerprint Verification Competition. In the last edition FVC 2004 [4], fingerprint images with lower image quality than those of previous campaigns were used. As a result, the error rates of best systems were found to be more than an order magnitude worse than those reported in earlier competitions using more controlled data. These dramatic effects have also been noticed in other recent comparative benchmark studies [5].

Several factors determine the quality of a fingerprint image: skin conditions (e.g. dryness, wetness, dirtiness, temporary or permanent cuts and bruises), sensor conditions (e.g. dirtiness, noise, size), user cooperation, etc. Some of these factors cannot be avoided and some of them vary along time. The purpose of this paper is to study the behavior of two common fingerprint matchers under image quality degradation by using a selection of quality measures available from the literature.

The paper is structured as follows. Sect. 2 introduces the two systems used in our study. Sect. 3 describes related work on the characterization of fingerprint image quality. Experimental setup and results are given in Sect. 4. Conclusions are finally drawn in Sect. 5.

## 2. FINGERPRINT VERIFICATION MATCHERS

We use both the minutia-based [3] and the ridge-based fingerprint matchers developed in the Biometrics Research Lab. at Universidad Autonoma de Madrid, Spain. In this paper we focus on fingerprint verification using these matchers. The system architecture of a fingerprint verification application is depicted in Fig. 1.

The minutiae matcher is based on the architecture presented in [6] with the modifications detailed in [3] and the references therein, resulting in a similarity measure $s_M$ based on dynamic programming. The output score is normalized into the [0,1] range by $\tanh(s_M/c_M)$, where $c_M$ is a normalization parameter chosen heuristically.

The ridge-based matcher consist of correlation of Gabor-filter energy responses in a squared grid as proposed in [7] with some modifications. No image enhancement is performed in the present work. Also, once the horizontal and vertical displacements maximizing the correlation are found, the original images are aligned and the Gabor-based features are recomputed before the final matching. The result is a dissimilarity measure $s_R$ based on Euclidean distance as in [7]. The output score is normalized into a similarity

This work has been supported by Spanish MCYT TIC2003-08382-C05-01 and by European FP6 IST-2002-507634 Biosecure NoE projects. J. F.-A. and F. A.-F. also thank Consejeria de Educacion de la Comunidad de Madrid and Fondo Social Europeo for supporting their doctoral research.

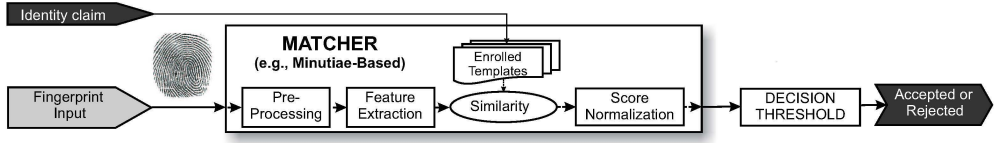

Fig. 1. System architecture of a fingerprint verification application.

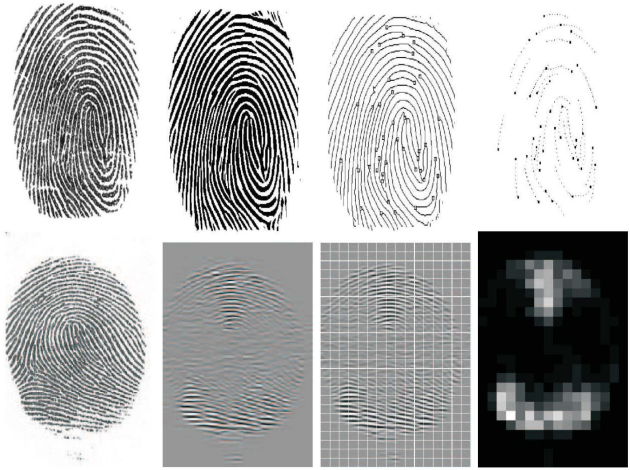

Fig. 2. Processing steps of the two matchers used.

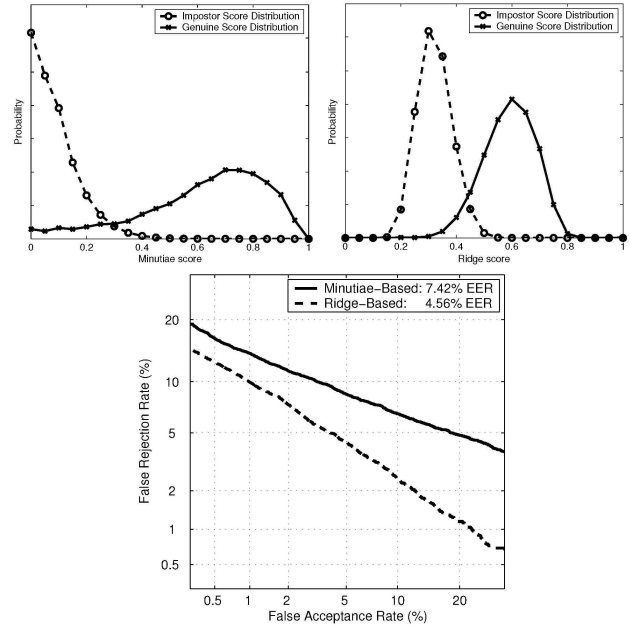

Fig. 3. Verification performance of the matchers.

measure in the [0,1] range by $\exp(-s_R/c_R)$ where $c_R$ is a normalization parameter chosen heuristically.

The pre-processing and feature extraction steps carried out in both matchers are shown in Fig. 2. Fig. 3 depicts the genuine and impostor score distributions as well as the verification performance of the two matchers for the whole database considered according to the experimental protocol defined in Sect. 4.1.

## 3. FINGERPRINT IMAGE QUALITY MEASURES

Fingerprint quality is usually defined as a measure of the clarity of ridges and valleys and the "extractability" of the features used for identification such as minutiae, core and delta points, etc. [8]. In good quality images, ridges and valleys flow smoothly in a locally constant direction. As a result, most of the fingerprint image quality measures already identified in the literature are based on operational procedures for computing local orientation coherence measures [9].

A taxonomy of existing approaches for fingerprint image quality computation is given in [10]. We can divide the existing approaches into $i$) those that use local features of the image; $ii$) those that use global features of the image; and $iii$) those that address the problem of quality assessment as a classification problem.

In the present work we use the following quality measures:

**Manual.** Each different fingerprint image is assigned a subjective quality measure $Q$ by a human expert from 0 (lowest quality) to 1 (highest quality) based on image factors like: incomplete fingerprint, smudge ridges or non uniform contrast, background noise, weak appearance of the ridge structure, significant breaks in the ridge structure, pores inside the ridges, etc. Fig. 4 shows four example images and their labeled quality according to this criterion.

**Local.** These methods usually divide the image into non-overlapped square blocks and extract features from each block. Blocks are then classified into groups of different quality. A local measure of quality is finally generated by averaging the quality information in the different blocks. In particular, we use the local measure recently proposed by Chen et al. [8] which measures the spatial coherence using the intensity gradient. A local quality score is computed by averaging

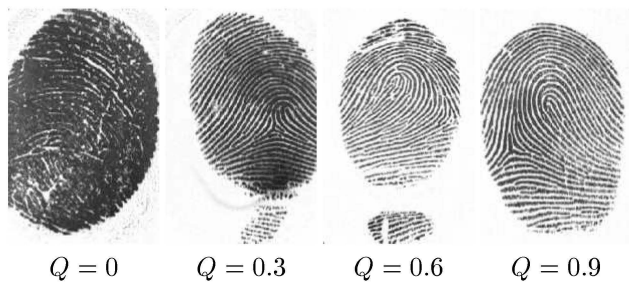

$Q = 0$   $Q = 0.3$   $Q = 0.6$   $Q = 0.9$

**Fig. 4**. Example fingerprint images.

the coherence of each block, weighted by its distance to the centroid of the foreground.

**Global.** Methods that rely on global features analyze the overall image and compute a global measure of quality based on the features extracted. In particular, we use the global measure proposed by Chen et al. [8] which is based on the Discrete Fourier Transform. The global quality index is a measure of the energy concentration in ring-shaped regions of the region of interest of the spectrum.

**NIST.** We use the publicly available NIST software for estimating fingerprint image quality [11]. The quality measure is defined as the degree of separation between the match and non-match distributions of a given fingerprint, which is predicted by using Neural Networks.

## 4. EXPERIMENTS

### 4.1. Database and experimental protocol

We use a subcorpus of the MCYT Biometric Database [12]. Data consist of 7500 fingerprint images from the 10 fingers of 75 contributors acquired with an optical sensor, model UareU from Digital Persona, with a resolution of 500 dpi and a size of 400 pixels height and 256 pixels width. We consider the different fingers as different users enrolled in the system, thus resulting in 750 users with 10 impressions per user. Some example images are shown in Fig. 4.

We use one impression per finger as template (with low control during the acquisition, see [12]). Genuine matchings are obtained comparing the template to the other 9 impressions available. Impostor matchings are obtained by comparing the template to one impression of all the other fingers. The total number of genuine and impostor matchings are therefore $750 \times 9$ and $750 \times 749$, respectively.

We further classify all the fingers in the database into five equal-sized disjoint quality groups, from I (low quality), to V (high quality), based on the quality measures described in Sect. 3, resulting in 150 fingers per group. Each quality group leads to $150 \times 9$ plus $150 \times 749$ matching scores. In order to classify the fingers in the five quality groups, a quality ranking of fingers is carried out. This ranking is based on the average quality of genuine matchings corresponding to the specific finger. The quality of a matching is defined as $\sqrt{Q_{\text{enrolled}} \cdot Q_{\text{input}}}$, where $Q_{\text{enrolled}}$ and $Q_{\text{input}}$ are the image qualities of the enrolled and input fingerprints respectively corresponding to the matching.

### 4.2. Results

In Fig. 5 we show verification performance results of the two matchers introduced in Sect. 2. Top row depicts the image quality distribution in the whole corpus (7500 images from 750 different fingers) according to the different criteria mentioned in Sect. 3 (from left to right): manual, local, global and NIST. Bottom row depicts verification performance results of the two matchers for different quality groups ($x$-axis of the different plots) and for the different quality criteria as in the top row.

By looking at the top row of Fig. 5 we first observe that image quality across the database is more os less evenly distributed in case of manual, local, and global quality measures. In case of the NIST quality measure, nearly 50% of the database is of the highest quality. The discrete nature of the manual and NIST quality measures (10 and 5 quality levels, respectively) is patent in the stepwise nature of these plots.

By observing the bottom row of Fig. 5 we observe that the ridge-based matcher is much more robust to image quality degradation than the minutiae-based matcher for all the considered quality measures. This is specially evident in the case of global quality, where the ridge-based matcher is found to be almost independent of the image quality. The largest drop in performance is for the minutiae-based matcher when considering the manual quality measure, from nearly 2.44% to 14.29% EER. As a result, we observe that the approach based on ridge information outperforms the minutiae-based approach in low image quality conditions for all quality measures considered. On the other hand, comparable performance between the two matchers is obtained on good quality images.

## 5. CONCLUSIONS

The effects of image quality degradation on the performance of two common approaches for fingerprint verification have been studied. It has been found that the approach based on ridge information outperforms the minutiae-based approach in low image quality conditions. Comparable performance is obtained on good quality images.

It must be emphasized that this evidence is based on particular implementations of established algorithms, and should not be taken as a general statement. Other implementations may lead to improved performance of any approach over the other in specific quality conditions. On the other hand, the robustness observed of the ridge-based approach as compared to the minutiae-based system has been observed in other works and can be consider quite general. Other study supporting this aim is the last Fingerprint Verification Competition in 2004 [4], where low quality images where used and leading systems used some kind of ridge information and not always minutiae [13].

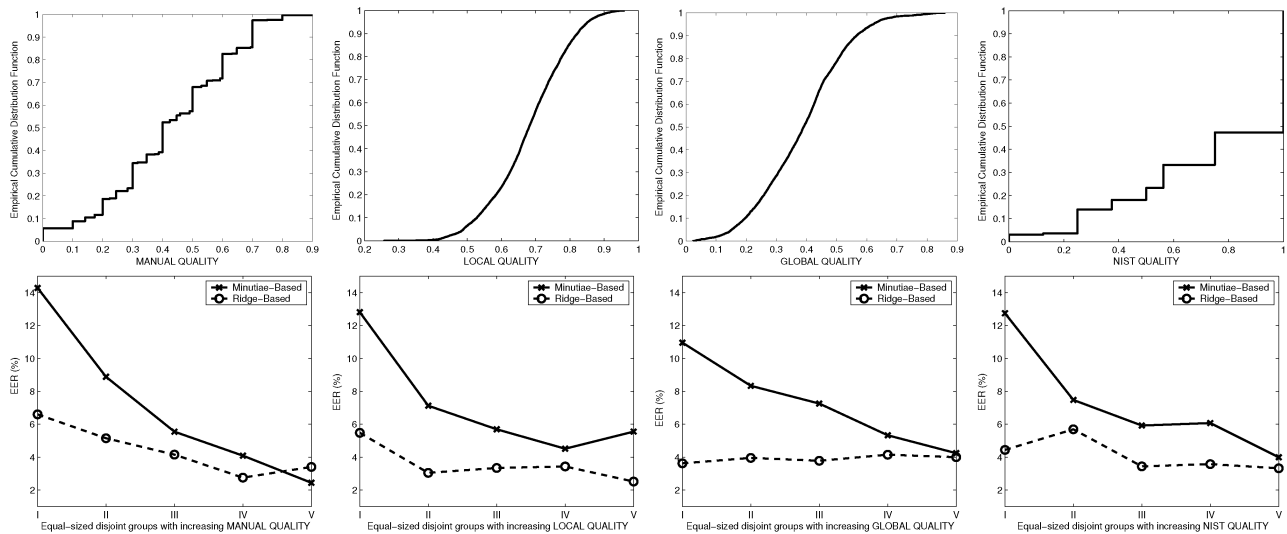

Fig. 5. Quality distributions (top) and verification results (bottom) for quality groups based on different quality criteria.

The different behavior of different approaches to fingerprint recognition under varying image quality, motivates us to conduct further research focused on multi-algorithm fingerprint verification schemes adapted to the image quality [14].

## 6. REFERENCES

[1] A. K. Jain, A. Ross, and S. Prabhakar, "An introduction to biometric recognition," IEEE Trans. CSVT, vol. 14, no. 1, pp. 4–20, 2004.

[2] D. Maltoni, D. Maio, A. K. Jain, and S. Prabhakar, Handbook of Fingerprint Recognition, Springer, 2003.

[3] D. Simon-Zorita, J. Ortega-Garcia, J. Fierrez-Aguilar, et al., "Image quality and position variability assessment in minutiae-based fingerprint verification," IEE Proc. VISP, vol. 150, no. 6, pp. 402–408, 2003.

[4] D. Maio, D. Maltoni, R. Cappelli, J. L. Wayman, and A. K. Jain, "FVC 2004: Third Fingerprint Verification Competition," Proc. ICBA, Springer LNCS, vol. 3072, pp. 1–7, 2004.

[5] C. Wilson et al., "FpVTE2003: Fingerprint Vendor Technology Evaluation 2003," NISTIR 7123, 2003.

[6] A. K. Jain, L. Hong, S. Pankanti, and R. Bolle, "An identity authentication system using fingerprints," Proc. IEEE, vol. 85, no. 9, pp. 1365–1388, 1997.

[7] A. Ross, J. Reisman, and A. K. Jain, "Fingerprint matching using feature space correlation," Proc. BioAW, Springer LNCS, vol. 2359, pp. 48–57, 2002.

[8] Y. Chen, S. Dass, and A. K. Jain, "Fingerprint quality indices for predicting authentication performance," Proc. AVBPA, Springer LNCS, vol. 3546, pp. 160–170, 2005.

[9] J. Bigun, G. H. Granlund, and J. Wiklund, "Multidimensional orientation estimation with applications to texture analysis and optical flow," IEEE Trans. PAMI, vol. 13, no. 8, pp. 775–790, 1991.

[10] F. Alonso-Fernandez, J. Fierrez-Aguilar, and J. Ortega-Garcia, "A review of schemes for fingerprint image quality computation," Proc. 3rd COST 275 Workshop, Biometrics on the Internet, European Commission, 2005. (to appear)

[11] E. Tabassi, C. Wilson, and C. Watson, "Fingerprint image quality," NIST Research Report NISTIR 7151, August 2004.

[12] J. Ortega-Garcia, J. Fierrez-Aguilar, D. Simon, J. Gonzalez, M. Faundez-Zanuy, V. Espinosa, A. Satue, I. Hernaez, J.-J. Igarza, C. Vivaracho, D. Escudero, and Q.-I. Moro, "MCYT baseline corpus: A bimodal biometric database," IEE Proc. VISP, vol. 150, no. 6, pp. 395–401, 2003.

[13] J. Fierrez-Aguilar, L. Nanni, J. Ortega-Garcia, R. Cappelli, D. Maltoni, "Combining multiple matchers for fingerprint verification: a case study in FVC2004," Proc. ICIAP, Springer LNCS, vol. 3617, pp. 1035-1042, 2005.

[14] J. Fierrez-Aguilar, Y. Chen, J. Ortega-Garcia, and A. K. Jain, "Incorporating image quality in multi-algorithm fingerprint verification," Proc. ICB, Springer LNCS, 2006. (to appear)